\documentclass[letterpaper, 10 pt, conference]{ieeeconf}  

\IEEEoverridecommandlockouts                              

\usepackage{times}
\usepackage{epsfig}
\usepackage{graphicx}
\usepackage{amsmath}
\usepackage{amssymb}
\usepackage{booktabs}

\usepackage[utf8]{inputenc} 
\usepackage[T1]{fontenc}    
\usepackage{url}            
\usepackage{booktabs}       
\usepackage{amsfonts}       
\usepackage{nicefrac}       
\usepackage{microtype}      
\usepackage{xcolor}         

\usepackage{times}
\usepackage{epsfig}
\usepackage{graphicx}
\usepackage{amsmath}
\usepackage{amssymb}
\usepackage{multirow}
\usepackage{booktabs}
\usepackage{forest}
\newcommand{\sysname}{Logic-RAG}

\usepackage{algorithm2e}

\usepackage[pagebackref,breaklinks,colorlinks]{hyperref}

\usepackage[capitalize]{cleveref}
\crefname{section}{Sec.}{Secs.}
\Crefname{section}{Section}{Sections}
\Crefname{table}{Table}{Tables}
\crefname{table}{Tab.}{Tabs.}

\definecolor{materialteal}{HTML}{009688}
\definecolor{materialpurple}{HTML}{9C27B0}
\definecolor{materialindigo}{HTML}{3F51B5}
\definecolor{materialblue}{HTML}{2196F3}
\definecolor{materialcyan}{HTML}{00BCD4}
\definecolor{materialteal}{HTML}{009688}
\definecolor{materialgreen}{HTML}{4CAF50}
\definecolor{materiallime}{HTML}{CDDC39}
\definecolor{materialamber}{HTML}{FFC107}
\definecolor{materialbrown}{HTML}{795548}
\definecolor{materialred}{HTML}{FF4436}
\definecolor{materialorange}{HTML}{FF5722}

\def \billahdebug{}

\ifx \billahdebug \undefined

\else

\fi

\usepackage{array}

\newcommand{\fColOf}[1]{\textit{ColorOf($#1$)}}
\newcommand{\fTypeOf}[1]{\textit{TypeOf($#1$)}}

\newcommand{\pRoad}[1]{\textit{Road($#1$)}}
\newcommand{\pLaneMarking}[1]{\textit{LaneMarking($#1$)}}
\newcommand{\pTrafficSign}[1]{\textit{TrafficSign($#1$)}}
\newcommand{\pSidewalk}[1]{\textit{Sidewalk($#1$)}}
\newcommand{\pFence}[1]{\textit{Fence($#1$)}}
\newcommand{\pPole}[1]{\textit{Pole($#1$)}}
\newcommand{\pWall}[1]{\textit{Wall($#1$)}}
\newcommand{\pBuilding}[1]{\textit{Building($#1$)}}
\newcommand{\pVegetation}[1]{\textit{Vegetation($#1$)}}
\newcommand{\pVehicle}[1]{\textit{Vehicle($#1$)}}
\newcommand{\pPedestrian}[1]{\textit{Pedestrian($#1$)}}
\newcommand{\pOther}[1]{\textit{Other($#1$)}}

\newcommand{\pAppears}[1]{\textit{Appears($#1$)}}
\newcommand{\pDisappears}[1]{\textit{Disappears($#1$)}}
\newcommand{\pMoves}[1]{\textit{Moves($#1$)}}
\newcommand{\pSpeedUp}[1]{\textit{SpeedUp($#1$)}}
\newcommand{\pSpeedDown}[1]{\textit{SpeedDown($#1$)}}
\newcommand{\pCloseToCamera}[1]{\textit{CloseToCamera($#1$)}}
\newcommand{\pAtRight}[1]{\textit{AtRight($#1$)}}
\newcommand{\pAtLeft}[1]{\textit{AtLeft($#1$)}}
\newcommand{\pAtCenter}[1]{\textit{AtCenter($#1$)}}

\newcommand{\pDistanceDecreases}[2]{\textit{DistanceDecreases($#1$, $#2$)}}
\newcommand{\pDistanceIncreases}[2]{\textit{DistanceIncreases($#1$, $#2$)}}
\newcommand{\pDistanceDecreasesToZero}[2]{\textit{DistanceDecreasesToZero($#1$, $#2$)}}
\newcommand{\pOn}[2]{\textit{On($#1$, $#2$)}}

\newcommand{\pGettingCloser}[2]{\textit{GettingCloser($#1$, $#2$)}}

\newcommand{\pCollide}[2]{\textit{Collide($#1$, $#2$)}}

\newcommand{\pWalk}[1]{\textit{Walk($#1$)}}
\newcommand{\pStopped}[1]{\textit{Stopped($#1$)}}
\newcommand{\pStand}[1]{\textit{Stand($#1$)}}
\newcommand{\pAccelerate}[1]{\textit{Accelerate($#1$)}}

\newcommand{\pConstantSpeed}[1]{\textit{ConstantSpeed($#1$)}}
\newcommand{\pIncreasePace}[1]{\textit{IncreasePace($#1$)}}

\newcommand{\pFixedPace}[1]{\textit{FixedPace($#1$)}}

\title{\LARGE \bf Logic-RAG: Augmenting Large Multimodal Models with Visual-Spatial Knowledge for Road Scene Understanding}

\author{Imran Kabir$^{1}$, Md Alimoor Reza$^{2}$, and Syed Billah$^{1}$
\thanks{$^{1}$College of Information Sciences and Technology, Pennsylvania State University, State College, PA 16801, USA
        {\tt\small \{ibk5106, skb5969\}@psu.edu}}%
\thanks{$^{2}$Department of Mathematics and Computer Science, Drake University, Des Moines, IA 50311, USA
        {\tt\small md.reza@drake.edu}}%
}

\begin{document}
\maketitle
\thispagestyle{empty}
\pagestyle{empty}

\begin{abstract}
Large multimodal models (LMMs) are increasingly integrated into autonomous driving systems for user interaction. However, their limitations in fine-grained spatial reasoning pose challenges for system interpretability and user trust. We introduce Logic-RAG, a novel Retrieval-Augmented Generation (RAG) framework that improves LMMs' spatial understanding in driving scenarios.  
Logic-RAG constructs a dynamic knowledge base (KB) about object-object relationships in first-order logic (FOL) using a perception module, a query-to-logic embedder, and a logical inference engine.  
We evaluated Logic-RAG on visual-spatial queries using both synthetic and real-world driving videos. When using popular LMMs (GPT-4V, Claude 3.5) as proxies for an autonomous driving system, these models achieved only 55\% accuracy on synthetic driving scenes and under 75\% on real-world driving scenes. Augmenting them with Logic-RAG increased their accuracies to over 80\% and 90\%, respectively. An ablation study showed that even without logical inference, the fact-based context constructed by Logic-RAG alone improved accuracy by 15\%.  
Logic-RAG is extensible: it allows seamless replacement of individual components with improved versions and enables domain experts to compose new knowledge in both FOL and natural language. In sum, Logic-RAG addresses critical spatial reasoning deficiencies in LMMs for autonomous driving applications.  
Code and data are available at: \url{https://github.com/Imran2205/LogicRAG}.  
\end{abstract}

\section{Introduction}
End-to-end AI systems such as autonomous vehicles are becoming mainstream, yet their black-box nature makes their decisions difficult to interpret~\cite{omeiza2021explanations}. 
These systems typically communicate through either visualization interfaces or natural language interactions powered by large multimodal models (LMMs)~\cite{yang2023survey, zhou2023vision, xu2024drivegpt4}. While visualizations show system state, they offer limited user interaction. LMM interfaces, in contrast, enable users to converse about and reason through system decisions, potentially improving interpretability~\cite{chen2023driving}.

\begin{figure*}[!t]
    \centering
    \includegraphics[width=0.99\textwidth]{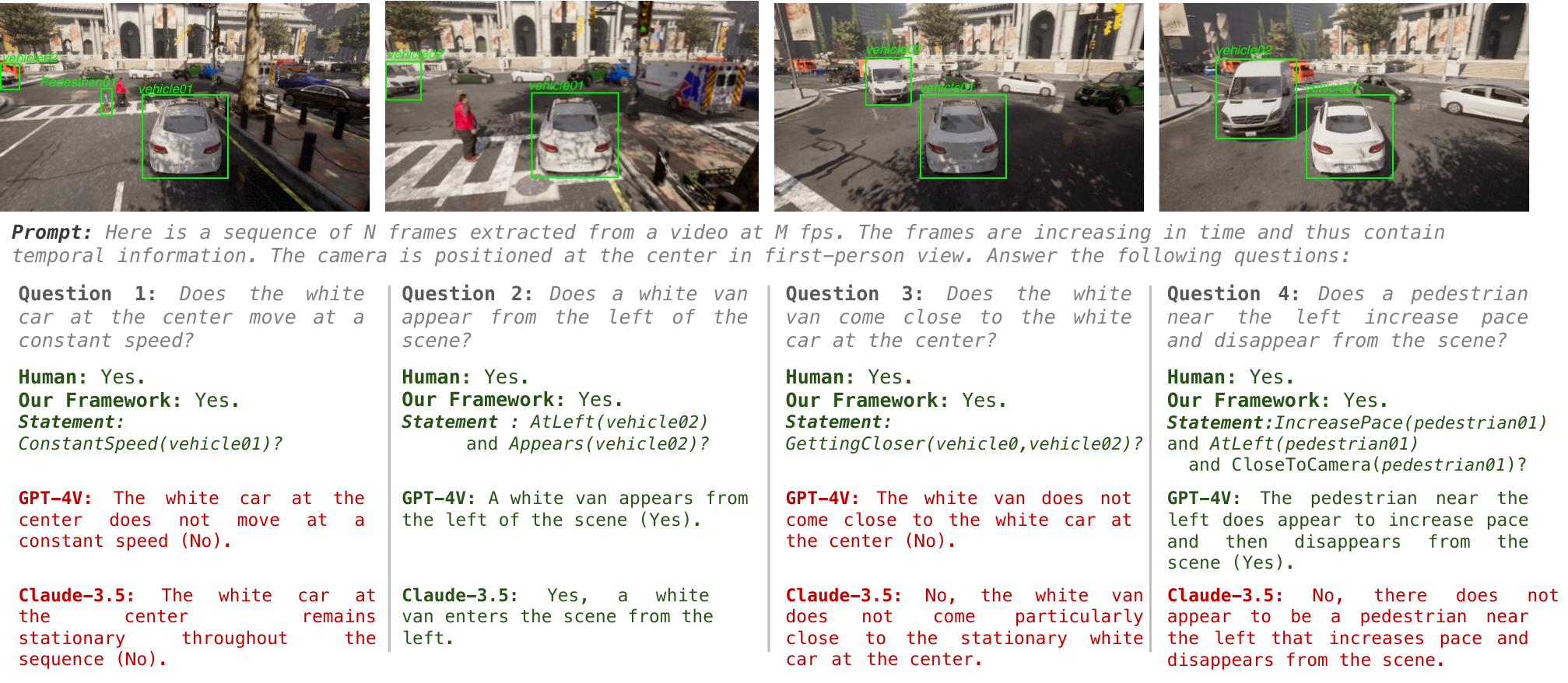}
    \caption{An illustration of the limitations of current commercial LMM models in visual-spatial reasoning (VSR). (Top) A sample of 4 consecutive frames from a synthetic driving video where the VSR task will be performed. (Middle) The original prompts and four representative questions. The values of N and M are 10 for GPT-4V and 5 for Claude-3.5, respectively, due to their current limits. (Bottom) Four responses from: 1) human oracle (accuracy: 100\%), 2) \sysname{}, 3) GPT-4V, and 4) Claude-3.5.
    The correct answers are colored green, and the incorrect answers are colored red. Note that \sysname{}'s accuracy is higher than black-box LMMs, and its response is generated by logical inference in our FOL system.
}
    \vspace{-15pt}
    \label{fig:application_events_compose}
\end{figure*}

However, LMMs struggle with fine-grained visual-spatial reasoning (VSR) -- they fail to accurately represent relationships between objects (e.g., \text{in front of}, \text{behind})~\cite{liu2023visualspatial, kamath2023s}, confuse nuanced action relationships (e.g., distinguishing between  \text{\textit{vehicle01} chased \textit{vehicle02}} vs. \text{\textit{vehicle02} chased \textit{vehicle01}})~\cite{thrush2022winoground}, and exhibit weak logical reasoning~\cite{logic_lm}. Moreover, LMMs can generate content that they do not fully understand~\cite{west2023generative} and are prone to hallucination~\cite{banerjee2024llms}.
Fig.\ref{fig:application_events_compose} demonstrates these limitations, showing how popular LMMs (GPT-4V~\cite{gpt4} and Claude 3.5~\cite{claude3}) struggle with even simple VSR questions like \textit{``Does the white car at the center move at a constant speed?''}. Such deficiencies hinder the integration of LMM interfaces into autonomous driving systems and undermine users' trust.

\begin{figure}[!t]
  \centering
  \includegraphics[width=0.95\linewidth]{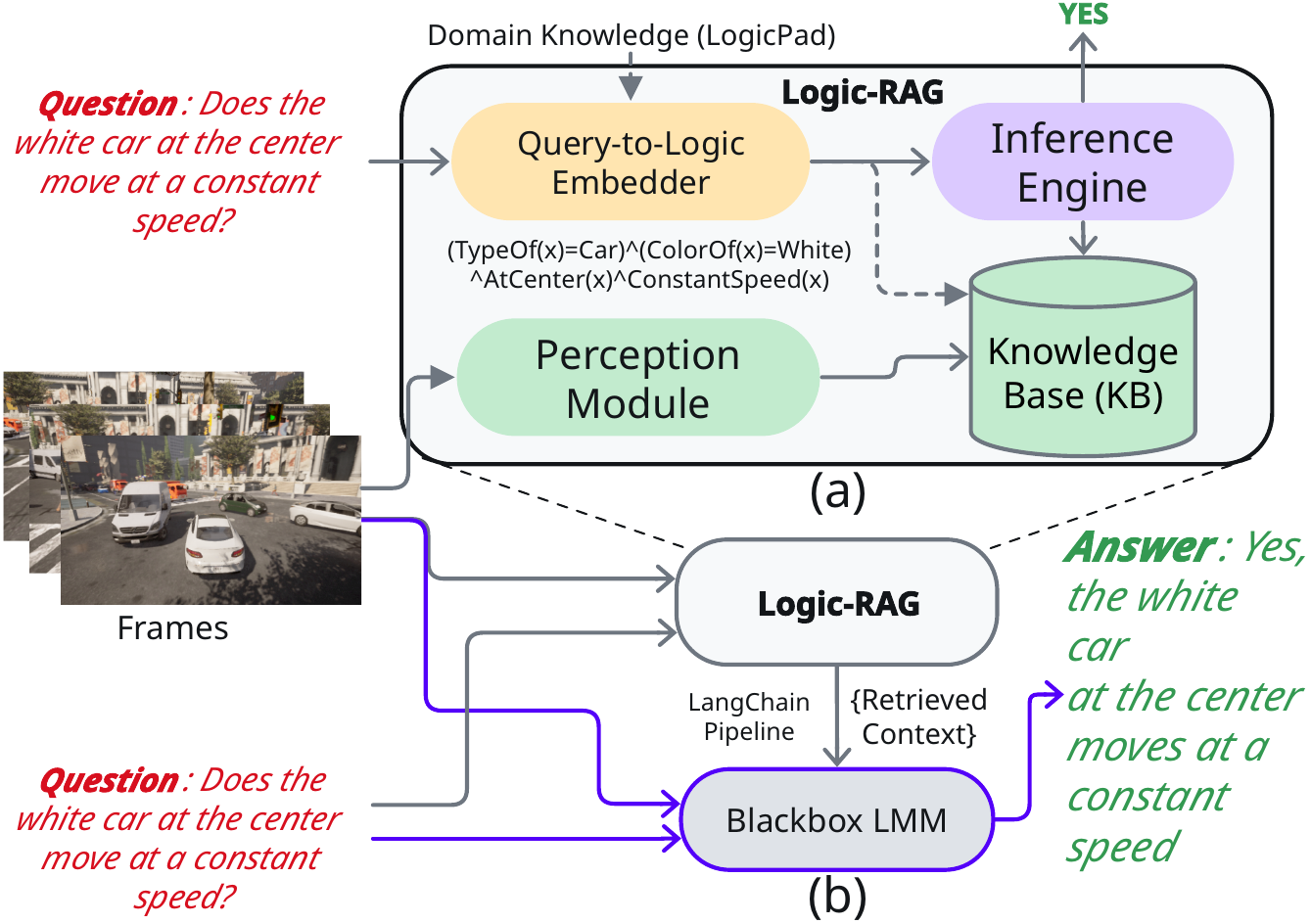}
  \caption{(a) Components of the Logic-RAG framework. It takes $N$ frames and visual-spatial reasoning questions as inputs. The perception module analyzes the frames to generate facts about properties and visual-spatial relationships of objects in the scene and constructs a knowledge base (KB) in the form of First-Order Logic (FOL). The Query-to-Logic embedder parses the natural language question into an FOL query predicate, which is then passed to the Inference Engine that performs the query resolution.
(b) The integration of \sysname{} into a black box LMM, which receives the inference output of our framework while generating the response.
}
\vspace{-15pt}
  \label{fig:teaser}
\end{figure}

Some of these limitations can be addressed by retrieval-augmented generation (RAG) techniques~\cite{lewis2020retrieval}, which provide LMMs with relevant information retrieved from vector storage as ``context'' to reduce factual errors in their responses.

Building on this concept, we present {\sc \textbf{\sysname{}}}, a novel RAG framework that constructs a knowledge base (KB) of facts about road scene objects and their relationships using first-order logic (FOL).
It comprises four key components (Fig.~\ref{fig:teaser}.a): 1) a \text{Perception} module, 2) a \text{Query-to-Logic Embedder}, 3) the \text{Knowledge Base} (KB), and 4) an \text{Inference Engine} that performs query resolution by applying inference rules to derive new facts until a contradiction is found or the desired query is proven. The facts retrieved from the inference are fed into LMMs for augmentation (Fig.~\ref{fig:teaser}.b), which enhances their spatial reasoning abilities in dynamic scenes while generating responses.
Like any expert system, human experts provide a small set of declarative rules and facts in FOL to bootstrap the framework~\cite{lu2018robot}.

We evaluated \sysname{} on visual-spatial relationship questions using both synthetic (Fig.~\ref{fig:application_events_compose}) and real-world (KITTI~\cite{geiger2012we}) driving videos.
We used popular commercial LMMs as proxies for black-box autonomous driving systems. On synthetic data, our results show that while these LMMs achieve an average accuracy of around 55\% (baseline), augmenting them with \sysname{} increases accuracy by more than 25\%, exceeding 80\%.  
An ablation study revealed that augmenting LMMs with facts represented as template sentences (without the logical inference module) still improved accuracy by over 10\% compared to the baseline condition. On real-world data, \sysname{} achieved 91\% accuracy, outperforming the baseline LMMs by 17\%.

These findings strongly suggest that when LMMs learn from joint embedding of image-text pairs~\cite{gupta2022towards}, they may have bypassed traditional computer vision pipelines (e.g., semantic segmentation, depth estimation, multi-object tracking), leading to deficiencies in spatial reasoning and object-object relationship understanding~\cite{rajabi2023towards}. By recovering this missing information and augmenting LMMs with it, our framework significantly improves their accuracy in visual-spatial reasoning tasks.

\noindent In sum, we make the following \textbf{contributions}:
\begin{itemize}
\item \textbf{\sysname{}:}
We introduce the \sysname{} framework, which encodes detailed scene facts into a dynamic knowledge base using First-Order Logic.

\item \textbf{Extensibility and Adaptability:}
Our framework allows the incorporation of new domain knowledge from human experts, using existing predicates to tailor the framework to specific tasks and utility.
\end{itemize}

\section{Background and Related Work}
\subsection{Logical Reasoning using First-Order Logic}
\label{sec:fol}
First-order logic (FOL) is an expressive formal system for representing and reasoning about objects, their properties, and relationships.
FOL has been fundamental in AI for knowledge representation and inference~\cite{russell2016artificial}. Neuro-symbolic approaches like DeepProbLog~\cite{manhaeve2018deepproblog} and Neural Theorem Provers~\cite{rocktaschel2017end} combine deep learning with symbolic reasoning. 
LOGIC-LM~\cite{logic_lm} demonstrated the integration of large language models with symbolic solvers for enhanced logical reasoning.

\paragraph{Components of FOL} 
\label{sub_sec: fol_components}

FOL consists of \textit{predicates}, \textit{functions}, \textit{constants}, \textit{variables}, and \textit{quantifiers}. 
Predicates describe properties of objects or relationships between objects. They take objects as arguments and produce a proposition that is either \textit{true} or \textit{false}. 
Functions, like predicates, can take any number of arguments but always return a single value rather than a proposition. For instance, \fColOf{\text{vehicle01}} is a function that returns the color of the vehicle represented by \textit{vehicle01}.
Constants represent specific objects, while variables are placeholders. Quantifiers, such as the universal ($\forall$) and existential ($\exists$) quantifiers, enable reasoning about multiple objects.
For example, if \pVehicle{\text{vehicle01}} is a specific segment of an object type vehicle in a scene and \pVehicle{x} is a base predicate indicating that variable $x$ is a vehicle, then \textit{vehicle01} is a constant in FOL. 

\paragraph{FOL Statements and Operators}
FOL statements are constructed by combining predicates, logical connectives (e.g., conjunction $\land$, disjunction $\lor$, negation $\lnot$, equality $=$, and implication $\to$), and quantifiers. 
For example, the statement 
\(  \forall x, y : ( \pDistanceDecreases{x}{y} ) \to \pGettingCloser{x}{y} \)
indicates that for all objects $x$ and $y$, if the distance between objects $x$ and $y$ decreases in a set of consecutive video frames, then $x$ and $y$ are getting closer. 
The equality ($=$) operator determines whether two objects or object properties are identical.
For example, if \textit{vehicle01} and \textit{vehicle02} are constants representing two black vehicles, the statement $\fColOf{\text{vehicle01}} = \fColOf{\text{vehicle02}}$ is \textit{true}.

\subsection{Retrieval Augmented Generation (RAG)}
Recent advances in RAG have reduced the hallucination problem of LMMs by combining the strengths of language models with information retrieval systems~\cite{lewis2020retrieval}. 
This allows the model to generate content based on accurate and up-to-date information~\cite{banerjee2024llms}. 
RAG works as follows: $y = LLM(q, R(q))$, where $y$ is the output generated by the $LMM(.,.)$, $q$ is the input (in this case, the question or query), and $R(q)$ is the relevant information retrieved from an external KB.
In \sysname{}, we employ RAG using a formal method. Our KB is composed of FOL facts, and $R(q)$ is the result of the resolution of the query $q$.

\vspace{-5pt}
\section{Method}
We now describe the four key components of \sysname{} (Fig.~\ref{fig:teaser}): 
the \textit{Knowledge Base} (KB), the \textit{Perception} module, the \textit{Query-to-Logic Embedder}, and the \textit{Inference Engine}. 
\sysname{} can be integrated with any LMM via LangChain (www.langchain.com) interface, a popular RAG pipeline.

\begin{figure}[!t]
  \centering
  \includegraphics[width=0.65\linewidth]{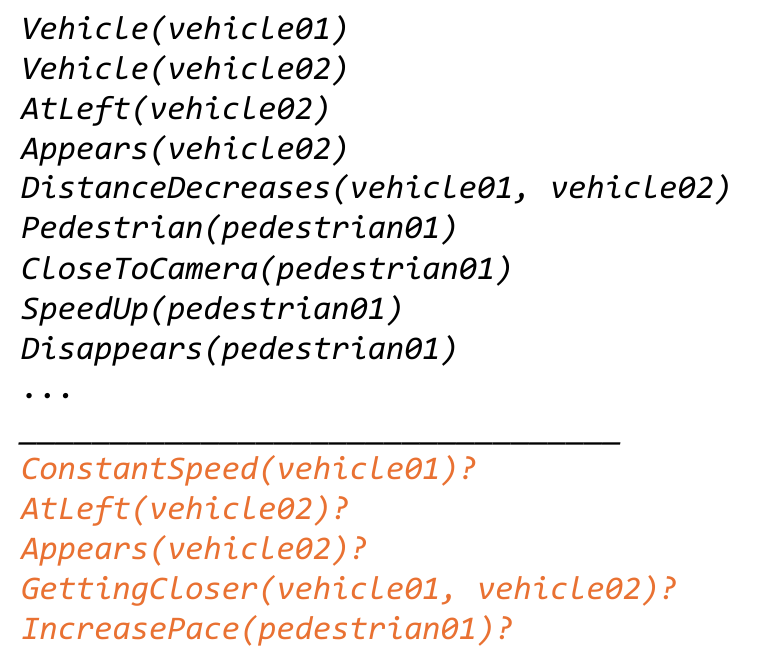}
  \caption{
  The predicates (in black) show a portion of KB that we construct using the output of the perception module for frames in Fig.\ref{fig:application_events_compose}. Text in orange shows the queries that we make in the KB to learn the facts. For instance, for the frames in Fig.\ref{fig:application_events_compose}, \pConstantSpeed{\text{vehicle01}} is \texttt{true}.}
  \vspace{-15pt}
  \label{fig:facts}
\end{figure}

\subsection{Knowledge Base (KB)}
Our Knowledge Base (KB) is comprised of base predicates, functions, and statements that represent object properties, spatial relationships, and events in driving contexts (Table~\ref{table:fol_kb}). The KB operates in a sliding window of $N$ consecutive video frames ($N=10$), where facts are generated by instantiating predicates with specific instances. For example, substituting \textit{vehicle01} for $x$ in \pVehicle{x} produces the fact \pVehicle{\text{vehicle01}}.
The KB updates dynamically as the sliding window progresses, with our multistage perception module (\S\ref{sec:perception}) continuously recognizing and incorporating new facts (see rows $2^{nd}$ and $3^{rd}$ in Table~\ref{table:fol_kb}). 
Fig.~\ref{fig:facts} shows some facts generated for the frames in Fig.~\ref{fig:application_events_compose}.

Importantly, our KB is extensible; it allows domain experts to compose predicates through \textit{LogicPad}, a text file written in YAML format that our system dynamically loads and interprets.
The capabilities of the perception module are exposed in LogicPad as atomic predicates and functions, similar to library methods in traditional programming languages.
Experts create new compound predicates (e.g., \pCollide{x}{y}) by combining atomic predicates and functions in FOL (see the \textit{Derived} rows in Table~\ref{table:fol_kb}). 
Alternatively, they can create compound predicates using natural language that our Query-to-Logic Embedder (\S\ref{sec:q2l}) translates into FOL.
As new capabilities are added in the perception module---either by swapping internal models or adding new ones---these capabilities automatically appear as new atomic predicates (or functions) in LogicPad. 
All predicates are available in the project \href{https://github.com/Imran2205/LogicRAG}{repository}.

\subsection{Perception Module}
\label{sec:perception}
\begin{figure}[!t]
\centering
  \includegraphics[width=0.98\linewidth]{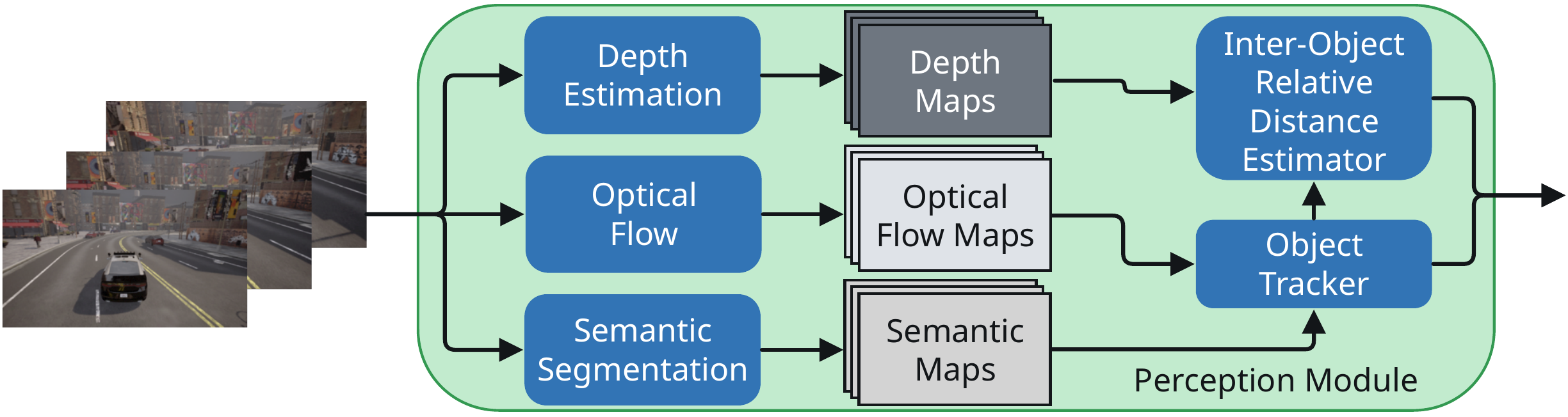}
  \caption{Internal block diagram of our perception module. It takes video frames as input and generates semantic, depth, and optical flow maps, which are then utilized to track object instances and estimate relative distances.}
  \vspace{-15pt}
  \label{fig:pipeline}
\end{figure}

Given an input driving video, we first generate dense prediction tasks for each video frame, which include (i) semantic segmentation, (ii) depth estimation, and (iii) optical flow prediction (see Fig.~\ref{fig:pipeline}).
We then track each semantic object using an object tracking model. Simultaneously, we compute the relative distance between objects within two adjacent frames inside the sliding window. These distances enable spatial reasoning among objects in the scene, which we subsequently use in our FOL predicates, such as \pGettingCloser{x}{y}, as discussed in \S\ref{sec:fol}.

\subsubsection{Dense Prediction Modules}
\label{sec:dense}
\noindent
Let us denote the input video comprising $M$ image frames. We also define a sliding window of $N$ consecutive frames, i.e., $\mathbf{I} = {I_1, I_2, ..., I_N}$. 
For each frame $I_i$, we need to identify various foreground and background objects as well as their semantic instances.
Furthermore, to describe inter-object relationships and the motion dynamics of different objects over multiple frames, we need to identify 3D spatial relationships between these objects, particularly depth and motion flow.
We describe these steps in the following sections.

\paragraph{Semantic Segmentation}
We start by densely predicting semantic labels for every pixel in an image frame $I_i$. This involves pixel-wise labeling of foreground objects, such as \textit{pedestrian} and \textit{vehicle}, as well as background objects, including \textit{sidewalk}, \textit{vegetation}, and \textit{building}.
Let us assume that there are $L$ semantic categories $\{c_1, c_2, ..., c_L\}$ in an image frame $I_i$, and each category $c_j$ can have $K_i$ instances, represented as $\{ s_{j1}^i, s_{j2}^i, \ldots, s_{jK_i}^i\}$.
Each semantic category is an atomic predicate in our framework (e.g., \pPedestrian{x}, \pVehicle{x}).
Instances of each category can then be encoded as \textit{constants} in FOL. For instance, we denote two instances obtained in the current image frame $I_i$ as \textit{pedestrian01} and \textit{vehicle01}.
These instances serve as \textit{constants} that are bound to \textit{atomic predicates}, in this case, \pPedestrian{\text{pedestrian01}} and \pVehicle{\text{vehicle01}}. 
The complete list of \textit{atomic predicates} used in our experiments is shown in Table~\ref{table:fol_kb}.
For semantic segmentation, we employed HRNet-v2~\cite{wang2020deep2} model, which has demonstrated effective generalization across multiple datasets trained from a composite one~\cite{mseg_cvpr20}.
Furthermore, to determine the color (e.g., `black', `blue', or 'white') and type (e.g., `car', `bus', or `SUV') of \textit{vehicle01}, the values of the functions \fColOf{\text{vehicle01}} and \fTypeOf{\text{vehicle01}} are predicted using a learned classifier.

\paragraph{Optical Flow}
To approximate the scene motion between two adjacent image frames in the video, we compute the optical flow for each pixel. These computed optical flows are then utilized to track important semantic instance segments, enabling us to understand the movement of objects within the scene.
We employ the state-of-the-art pretrained Rigid Mask model~\cite{yang2021rigidmask}, which excels at estimating optical flow and scene motion.

\paragraph{Depth Estimation}
Alongside semantic segmentation and optical flow, we estimate dense depth from an input image. Our depth estimation module provides per-pixel 3D depth values, which are crucial for identifying 3D spatial relationships between various semantic instance segments. These relationships encompass both intra- and inter-segment connections that evolve over time, as discussed in \S\ref{sec:inter_object_relative_distance}. To estimate the depth of various segments in each frame $I_i$, we employ the recent state-of-the-art depth estimation model called PixelFormer~\cite{Agarwal_2023_WACV}, which has shown impressive results in predicting depth information from a single image.

\subsubsection{Object Tracking for Semantic Instances}
\label{sec:tracking}
\noindent
We track the movement trajectory of each instance  $s_{jk}^i$ belonging to all semantic categories across all image frames within our specified window by framing it as a bipartite matching problem. These trajectories are implicitly used to describe single-object and multi-object interactions, which we will describe in detail in \S\ref{sec:inter_object_relative_distance}.
To construct the bipartite graph, we treat each instance as a vertex, and edges are the dense connections between the vertices of two frames.
We assign edge weights using two factors: an optical flow-based factor derived from accumulated flow vectors of instance $s_{jk}^i$  at position $s_{jk}^{i+1}$~\cite{reza2017label}, and a bounding box-based factor using the intersection over union (IoU) of boxes encompassing two vertices connected by an edge. 
We address the challenges from under-segmentation and occlusion  by inserting hypothetical vertices as described in \cite{ZamirECCV12}.

This process provides trajectories of individual instances within the specific sliding window.
Based on these trajectories, we encode predicates, such as \pAppears{x} and \pDisappears{x}, to represent the appearance and disappearance of instances.
Consider an instance \textit{vehicle01} of the category \pVehicle{x}. If \textit{vehicle01} is first detected in the current sliding window, the predicate \pAppears{\text{vehicle01}} becomes \textit{true} within that window, indicating that the instance has just appeared.
However, if \textit{vehicle01} was already present at the beginning of the current sliding window, \pAppears{\text{vehicle01}} will be false (i.e., $\neg$\pAppears{\text{vehicle01}}), indicating that it existed prior to the current window.

\begin{table*}[t]
\centering
\resizebox{15.5cm}{!}{
\begin{tabular}{p{1.0cm}cc}
\toprule

 \textbf{Submodule}
 & %
 &  \textbf{Atomic Predicates and Statements}\\
\midrule
\textbf{Semantic} & \multicolumn{2}{c}{
\pRoad{x}, 
\pLaneMarking{x}, 
\pTrafficSign{x}, 
\pSidewalk{x},
\pFence{x},
\pPole{x},
\pWall{x}, 
\pBuilding{x}, 
\pVegetation{x},
\pVehicle{x}, 
} \\

\textbf{Segmentation} &
\multicolumn{2}{c}{
\pPedestrian{x}, 
\pOther{x}
} \\

\midrule
\textbf{Tracker} & \multicolumn{2}{c}{
\pAppears{x}, 
\pDisappears{x}, 
\pMoves{x},  
\pSpeedUp{x}, 
\pSpeedDown{x}, 
\pCloseToCamera{x},
\pAtRight{x} , 
\pAtLeft{x}, 
\pAtCenter{x},
}
\\
& \multicolumn{2}{c}{
\pDistanceIncreases{x}{y},
\pDistanceDecreases{x}{y},
\pDistanceDecreasesToZero{x}{y}
\pOn{x}{y}
}
\\ 

\midrule
\textbf{Functions} &  & 
\fColOf{x}, \fTypeOf{x} \\

\midrule

& Stopped & 
$\forall x: ( \neg \pMoves{x} ) \to \pStopped{x}$
\\

\cmidrule{2-3}

& Walk & 
$\forall x: (\pPedestrian{x} \land \pMoves{x} ) \to \pWalk{x}$ \\
\cmidrule{2-3}

& Stand & 
$ \forall x: (\pPedestrian{x} \land \neg \pMoves{x}) \to \pStand{x} $ \\

\cmidrule{2-3}

\textbf{Derived} & Accelerate & 
$ \forall x: (\pVehicle{x} \land \pSpeedUp{x}) \to \pAccelerate{x} $  \\

\cmidrule{2-3}
& Constant Speed & 
$ \forall x: (\pVehicle{x} \land \neg\pSpeedUp{x} \land \neg\pSpeedDown{x} ) \to \pConstantSpeed{x} $  \\

\cmidrule{2-3}
& Increase Pace & 
$ \forall x: (\pPedestrian{x} \land \pSpeedUp{x}) \to \pIncreasePace{x} $  \\

\cmidrule{2-3}
& Fixed Pace & 
$ \forall x: (\pPedestrian{x} \land \neg\pSpeedUp{x} \land \neg\pSpeedDown{x}) \to \pFixedPace{x} $  \\

\cmidrule{2-3}

& Getting closer 
& 

 $  \forall x, y : (\pDistanceDecreases{x}{y}) \to \pGettingCloser{x}{y} $
 \\

 \cmidrule{2-3}

& Collide
& 
 $  \forall x, y : (\pDistanceDecreases{x}{y} \land \pDistanceDecreasesToZero{x}{y}) \to \pCollide{x}{y} $ 
 \\

 \cmidrule{2-3}

& $\dots$
& 
 $\dots$ \\

\bottomrule
\end{tabular}
}
\caption{A partial list of base predicates, functions, and statements representing object properties, spatial relationships, and events.}
\vspace{-25pt}
\label{table:fol_kb}
\end{table*}

\subsubsection{Inter-Object Relative Distance Estimator}
\label{sec:inter_object_relative_distance}

Using instance trajectories, we also determine the spatial displacement of inter-instance and intra-instance segments between two consecutive frames $I_i$ and $I_{i+1}$. These displacements are used to encode predicates such as \pMoves{x}, \pSpeedUp{x}, \pSpeedDown{x}, \pAtLeft{x}, \pCloseToCamera{x}, and \pDistanceIncreases{x}{y}, which represent various spatial relationships and movements of objects.

To calculate the spatial displacement, we first utilize the estimated depth map to obtain the 3D coordinates of each pixel in the current frame's camera coordinate space. 
Given a pixel with 2D image coordinates $(x^{j}, y^{j})$, we compute its corresponding 3D coordinates $(X^{j}, Y^{j}, Z^{j})$ using standard perspective projection: $X^{j} = \frac{(x^{j} - c_x) z^{j}}{f_{x}}$, $Y^{j} = \frac{(y^{j} - c_y) z^{j}}{f_{y}}$, and $Z^{j} = z^{j}$. Here, $z^{j}$ is the depth value for the $j^{th}$ pixel retrieved from the estimated depth map, $(c_x, c_y)$ is the optical center of the camera, and $(f_x,f_y)$ is the focal length of the camera.
For our visual-spatial reasoning, we consider the $X$-axis and $Z$-axis planes (bird's eye view) by discarding the $Y$-axis component along the height. This means that we only use the $(X, Z)$ coordinates when reasoning about object-object relationships and their spatial displacements.

To estimate the relative displacement of objects across frames, we need to establish a common world coordinate system origin for all frames within the window. Traditional approaches for this estimation include the SLAM~\cite{durrant_bailey_part1_RA06} or Structure from Motion (SfM)~\cite{furukawa2015multi} techniques. However, these techniques are computationally expensive.
As an alternative, we propose an approximation strategy that leverages the relative stability of static objects compared to dynamic objects in two consecutive frames. Our approach assumes that the locations of static objects (e.g., trees, buildings, and poles) remain nearly constant between two consecutive frames, while dynamic objects (e.g., vehicles, persons) move more significantly.
We feed the locations of static objects from two consecutive frames into a \textit{multilateration}~\cite{multilateration1973} algorithm, which yields a temporary origin that remains mostly constant between the two frames. This temporary origin serves as a reference point for estimating the relative displacement of moving objects between consecutive frames. This enables efficient computation of object trajectories and spatial relationships.

We use the displacement calculated from the temporary origin to derive the velocity and acceleration of moving objects. We then utilize these values to establish various predicates in our KB. For example, we evaluate the predicate \pMoves{\text{vehicle01}} as true if the displacement of the object \textit{vehicle01} within a sliding window exceeds 0. Similarly, we evaluate the predicate \pSpeedUp{\text{vehicle01}} as true if the velocity of \textit{vehicle01} at the end of a sliding window exceeds its initial velocity.

\subsection{Query-to-Logic Embedder}
\label{sec:q2l}
We employ an open-source large language model (e.g., LLaMa 3~\cite{dubey2024llama} or Phi-3~\cite{abdin2024phi}) with few-shot prompting. Our prompt includes: a) Our predefined predicates; b) Pairs of natural language statements and their corresponding FOL predicates; and c) A few pairs of natural language questions and their corresponding FOL queries.
This concise yet expressive prompt enables the conversion of natural language queries into FOL predicates in conjunctive normal form (CNF), which we term (CNF) embedding.
For example, given the natural language question ``Does the white car at the center move at a constant speed?'', our Query-to-Logic Embedder generates the logic: $(\fTypeOf{x}=Car) \land (\fColOf{x}=White) \land \pAtCenter{x} \land \pConstantSpeed{x}$.
During our evaluation of 645 natural language questions, we observed no errors in the embedder's output, demonstrating its robustness. The resulting FOL query is then passed to the Inference Engine for processing.

\subsection{Inference Engine}
We built our inference system with two established tools: \textit{miniKanren}~\cite{willard2020minikanren}, a symbolic computation framework, and \textit{aima-python}~\cite{aimacode}, which implements FOL representation and inference based on Russell and Norvig's work~\cite{russell2016artificial}.
To derive new facts from our KB, we apply standard logical principles such as modus ponens and resolution. For example, to identify objects approaching a specific vehicle, we can query \pGettingCloser{\text{vehicle01}}{y}, which binds variable $y$ to all objects moving closer to \textit{vehicle01} within the current window.
Our inference is sound; it ensures that we derive only valid conclusions from the given premises. 

\section{Visual-Spatial Reasoning Experiments}
\begin{table*}[htbp]
    \centering
    \resizebox{17.5cm}{!}{
    \begin{tabular}{p{1.2cm}|ccc|ccccc|ccccc}
        \specialrule{1.2pt}{0pt}{0pt}
        \multirow{4}{*}{\begin{tabular}{@{}c@{}}Relationship\\  \end{tabular}} &
        \multirow{4}{*}{\begin{tabular}{@{}c@{}}Question\\ Type  \end{tabular}} &
        \multirow{4}{*}{\begin{tabular}{@{}c@{}}No. of\\ Questions  \end{tabular}} &
        \multirow{4}{*}{\begin{tabular}{@{}c@{}}\textbf{\sysname{}}\\ (Oracle) \\ Accuracy/$F_1$ \end{tabular}} &
        \multicolumn{5}{c}{Accuracy$\uparrow$} & 
        \multicolumn{5}{|c}{$F_1$$\uparrow$} \\
        \cline{5-9}
        \cline{10-14}
        & & & & 
        \multirow{3}{*}{\begin{tabular}{@{}c@{}}\textbf{LR}/ \\ \textbf{LMM-1} (w LR)/ \\ \textbf{LMM-2} (w LR) \\ \end{tabular}} & 
        \multirow{3}{*}{\begin{tabular}{@{}c@{}}\textbf{LMM-1} \\ (w LR$^\dagger$ ) \\ 
        \end{tabular}} & 
        \multirow{3}{*}{\begin{tabular}{@{}c@{}}\textbf{LMM-1} \\ (w/o LR) \\ 
        \end{tabular}} & 
        \multirow{3}{*}{\begin{tabular}{@{}c@{}}\textbf{LMM-2}  \\ (w LR$^\dagger$) \\  \end{tabular}} & 
        \multirow{3}{*}{\begin{tabular}{@{}c@{}}\textbf{LMM-2} \\ (w/o LR) \\ \end{tabular}} & 
        \multirow{3}{*}{\begin{tabular}{@{}c@{}}\textbf{LR}/ \\ \textbf{LMM-1} (w LR)/ \\ \textbf{LMM-2} (w LR) \\ \end{tabular}} & 
        \multirow{3}{*}{\begin{tabular}{@{}c@{}}\textbf{LMM-1}  \\ (w LR$^\dagger$)  \\  
        \end{tabular}} & 
        \multirow{3}{*}{\begin{tabular}{@{}c@{}}\textbf{LMM-1} \\  (w/o LR) \\ \end{tabular}} & 
        \multirow{3}{*}{\begin{tabular}{@{}c@{}}\textbf{LMM-2}  \\ (w LR$^\dagger$)  \end{tabular}} & 
        \multirow{3}{*}{\begin{tabular}{@{}c@{}}\textbf{LMM-2} \\ (w/o LR) \\  \end{tabular}} \\
        & & & & & & & & & & \\
        & & & & & & & & & & \\
        \hline
        \multirow{4}{*}{Unary} & \textbf{U1} & 193 & 1.0/1.0 & \textbf{0.82} & 0.70 & 0.78 & 0.77 & 0.74 & \textbf{0.90} & 0.82 & 0.87 & 0.87 & 0.85  \\
        \cline{2-14}
        & \textbf{U2} & 79 & 1.0/1.0 & \textbf{0.82} & 0.77 & 0.43 & 0.71 & 0.53 & \textbf{0.90} & 0.87 & 0.60 & 0.83 & 0.69  \\
        \cline{2-14}
        & \begin{tabular}{@{}l@{}} \textbf{U3} \end{tabular} & 129 & 1.0/1.0 & \textbf{0.84} & 0.74 & 0.35 & 0.68 & 0.51 & \textbf{0.87} & 0.79 & 0.44 & 0.75 & 0.60  \\
        \cline{2-14}
        & \begin{tabular}{@{}l@{}}\textbf{U4} \end{tabular} & 98 & 1.0/1.0 & \textbf{0.83} & 0.63 & 0.19 & 0.63 & 0.39 & \textbf{0.90} & 0.76 & 0.31 & 0.76 & 0.55   \\
        \specialrule{1.2pt}{0pt}{0pt}
        \multirow{2}{*}{Binary} & \begin{tabular}{@{}l@{}}\textbf{B1} \end{tabular} & 84 & 1.0/1.0 & \textbf{0.79} & 0.70 & 0.77 & 0.73 & 0.63 & \textbf{0.88} & 0.83 & 0.87 & 0.84 & 0.77   \\
        \cline{2-14}
        & \begin{tabular}{@{}l@{}}\textbf{B2} \end{tabular} & 62 & 1.0/1.0 & \textbf{0.77} & 0.58 & 0.42 & 0.56 & 0.45 & \textbf{0.87} & 0.73 & 0.59 & 0.72 & 0.62  \\
        \specialrule{1.2pt}{0pt}{0pt}
        \begin{tabular}{@{}l@{}}Aggregated \\ Result\end{tabular} & & 645 & 1.0/1.0 & \textbf{0.82} & 0.69 & 0.54 & 0.72 & 0.57 & \textbf{0.88} & 0.79 & 0.68 & 0.81 & 0.70   \\
        \specialrule{1.2pt}{0pt}{0pt}
    \end{tabular}
    }
    \caption{Accuracy and $F_1$ scores of \sysname{} and baselines for various VSR question types. $LR$ represents \sysname{}, while $LR^\dagger$ denotes \sysname{}'s KB exported as template sentences without logical inference.}
    \vspace{-25pt}
    \label{table:model_comparison_all_vid}
\end{table*}

We conducted the experiment by first creating a synthetic dataset with the CARLA~\cite{dosovitskiy2017carla} simulator, to ensure we have access to ground truth to evaluate our perception module.
We then composed questions from this dataset to assess the effectiveness of our framework against two state-of-the-art commercial models, LMM-1: GPT-4V~\cite{gpt4} and LMM 2: Claude 3.5~\cite{claude3}. Lastly, we integrate our framework into these LMMs with and without logical inference and measure performance differences. We also conduct an experiment on KITTI object tracking dataset~\cite{geiger2012we} to evaluate the effectiveness of our framework on real-world scenarios.

\vspace{-7pt}
\subsection{Reasoning on Synthetic Dataset}
\paragraph{Evaluation Dataset}
\label{sec:experiment_eval_dataset}
We generated distinct environmental conditions in Carla Simulator by altering various properties, including \textit{Day}, \textit{Night}, \textit{Rain}, and \textit{Sunset}. We captured 5,256 video frames at 10 fps, divided into 525 sequences of 10 consecutive frames each. For each frame, we extracted the corresponding depth map, semantic segmentation map with 12 categories, and instance segmentation map.

\paragraph{Implementation Details of the Perception Module}
We fine-tuned several models in our perception module (Fig.~\ref{fig:pipeline}) using a separate training dataset of 9,706 synthetic RGB images with pixel-level annotations for semantic segmentation and depth. 
For the semantic segmentation submodule, we fine-tuned HRNetv2~\cite{wang2020deep2} using its official PyTorch implementation. We used the pre-trained model of Rigid Mask~\cite{rigidmasksceneflow} for the optical flow submodule without fine tuning. For the depth estimation submodule, we fine-tuned PixelFormer~\cite{Agarwal_2023_WACV} using its official implementation. We conducted all training and inference on a single Nvidia RTX A6000 GPU with 48GB memory. Note that input and output interfaces of these models are well-defined so that we can seamlessly replace a model with a better one within the perception module.

\paragraph{Evaluation Criteria} 
From the evaluation dataset, we selected 95 sequences of 10 consecutive frames and composed 3-9 visual-spatial reasoning (VSR) questions per sequence. These questions included object attributes, actions, object-object relationships, and events. We ensure question validity by focusing on objects present for at least one-third of the sliding window, referring to them by color and position. 
We composed all questions manually. All authors collaboratively analyzed each frame sequence to ensure questions were answerable from visual information in the frames, without introducing bias toward any specific model architecture. 
In total, we prepared 645 such questions.

We manually categorize questions into \textit{Unary} (about one object) and \textit{Binary} (involving two objects). Unary subcategories include \textit{object query} (\textbf{U1}), \textit{velocity} (\textbf{U2}), \textit{change in velocity} (\textbf{U3}), and object \textit{appearance/disappearance} (\textbf{U4}). The binary subcategories are \textit{ relative position} (\textbf{B1}) and \textit{relative distance change} (\textbf{B2}). 

For each frame sequence and question set, we generated responses under three conditions:
\begin{itemize}
\item[C1] \sysname{}, LMM-1, and LMM-2  in standalone.
\item[C2] LMM-1 and LMM-2 with \sysname{} via LangChain.
\item[C3] LMM-1 and LMM-2 provided with template sentences of facts from \sysname{} (ablation, no logical inference).
\end{itemize}
We manually verified all responses under each condition, compared results with the oracle, and reported accuracy and $F_1$ scores in Table~\ref{table:model_comparison_all_vid}.

\paragraph{Results: Condition C1}
When we populate the KB using ground truth for the perception module, \sysname{} unsurprisingly achieves 100\% accuracy. This suggests that our FOL predicates can express and infer the correct response. However, our current implementation of the perception module reaches 82\% average accuracy ($F_1$ score: 0.88). Although it outperforms standalone LMM-1 and LMM-2 in all question categories (overall accuracy: 0.54 and 0.57; overall $F_1$ score: 0.68 and 0.70, respectively),  it indicates room for improvement, which is feasible by replacing current models with newer, better ones.

\paragraph{Results: Condition C2}
LMM-1 and LMM-2 integrated with \sysname{} are the best-performing models that yields the same performance as the standalone \sysname{}. This is because LMMs' responses are conditioned by the inference output of \sysname{}.

\paragraph{Results: Condition C3}
LMMs with only the kB (in template sentences) of \sysname{} still outperform LLMs without it.    
For example, LMM-1 achieves an accuracy of 0.69 and an $F_1$ score of 0.79, while LMM-2 reaches 0.72 and 0.81, respectively. Compared to standalone LMMs, this represents a 15\% improvement in accuracy and an 11\% increase in $F_1$ score. The integration of template sentences significantly enhances LMMs' performance, particularly for complex reasoning questions. For \textbf{U3}, we observe $F_1$ score improvements of 0.35 and 0.15 for LMM-1 and LMM-2, respectively. Similarly, for \textbf{B2}, improvements of 0.14 and 0.10 are observed in the $F_1$ score for LMM-1 and LMM-2, respectively (Table~\ref{table:model_comparison_all_vid}).

\paragraph{Results: VSR question types}
LMM-1 and LMM-2 in standalone condition perform reasonably well in simple \textit{object queries} ($F_1$ scores: $0.87$ and $0.85$), but struggle with deeper spatial understanding tasks. Their performance decreases for \textit{ changes in velocity} (\textbf{U3}) ($F_1$ scores: $0.44$ and $0.60$) and \textit{relative distance change} (\textbf{B2}) ($F_1$ scores: $0.59$ and $0.62$), indicating limitations in complex spatial reasoning.

\subsection{Testing the Replacibility of Components within Perception Module and Performance on Real-World Driving Dataset}
In this experiment, we evaluated how easily one could replace various components in the perception module by substituting them with newer ones. Specifically, we substituted HRNetv2~\cite{wang2020deep2} with Mask2Former~\cite{cheng2021mask2former_} and Rigid Mask~\cite{rigidmasksceneflow} with CoTracker3~\cite{karaev2024cotracker3}. We used these models without fine-tuning or modifying other system components.

We used the test set of the KITTI object tracking dataset~\cite{geiger2012we}. It contained 29 real-world video sequences (84 to 1,175 frames each, totaling approximately 11,000 frames).
Similar to our previous experiment, we cut these sequences into 10-frame clips, yielding 1,100 clips total. From this collection, we randomly selected 19 clips and manually created 100 visual-spatial reasoning questions with corresponding ground-truth answers.
We then generated responses under condition C1 to compare our reconfigured system's real-world performance directly against baseline methods (LMM-1 and LMM-2) operating in standalone mode.

\begin{table}[htbp]
    \centering
    \resizebox{8.5cm}{!}{
    \begin{tabular}{p{3.5cm}|>{\centering\arraybackslash}p{2.5cm}|>{\centering\arraybackslash}p{2.5cm}}
        \specialrule{1.2pt}{0pt}{0pt}
        Method &
        \multicolumn{1}{c}{Accuracy$\uparrow$} & 
        \multicolumn{1}{|c}{$F_1$$\uparrow$} \\
        \hline
        \textbf{Logic-RAG} & \textbf{0.91} & \textbf{0.95} \\ 
        \textbf{LMM-1} (w/o LR) & 0.71 & 0.83 \\
        \textbf{LMM-2} (w/o LR) & 0.74 & 0.84 \\
        \specialrule{1.2pt}{0pt}{0pt}
    \end{tabular}
    }
    \caption{Accuracy and $F_1$ scores of \sysname{} with updated perception module and baselines for VSR questions on KITTI dataset.}
    \vspace{-20pt}
    \label{table:model_comparison_kitti}
\end{table}

\paragraph{Results}
\sysname{} with its updated perception module achieved 91\% accuracy ($F_1$ score: 0.95), outperforming both baseline methods. Compared to LMM-1 (accuracy: 71\%; $F_1$ score: 0.83) and LMM-2 (accuracy: 74\%; $F_1$ score: 0.84), our system demonstrated improvements of 20 and 17 percentage points respectively. Table~\ref{table:model_comparison_kitti} presents these comparative results in detail.

\vspace{-5pt}
\section{Conclusion}
We introduce \sysname{}, a first-order logic-based framework that enhances visual-spatial reasoning capabilities of popular LMMs through the retrieval-augmented generation mechanism. 
\sysname{} constructs a dynamic knowledge base about object-object relationships and spatial dynamics in driving scenes. This work addresses crucial spatial reasoning deficiencies in LMMs for autonomous driving applications, improving system interpretability and user trust. Future research could focus on extending the framework to more complex scenarios, incorporating temporal reasoning, and exploring real-time optimizations for practical deployment in autonomous vehicles.

\section{Acknowledgement}
This work was supported in part by the National Science Foundation under grant \#2326406.

\bibliographystyle{IEEEtran}
\bibliography{Bibliography, Bibliography2, Bibliography3, local}

\end{document}